**Title:** Generalizable fetal brain biometry via adaptive artificial intelligence

**Authors:** Yijin Li[#], BEng; Mingxuan Liu[#], BEng; Yi Liao[#], MD; Le Chang, BEng; Hongjia Yang, BEng; Kasidit Anmahapong, BMed; Xiaotian Hu, BEng; Ziang Wang, BMed; Yingqi Hao, BEng; Haoxiang Li, BEng; Yifei Chen, BEng; Zihan Li, BEng; Ziyu Li, PhD; Juncheng Zhu, MMed; Min Kang, MD; Yan Song, MD; Hua Lai, MD; Xiaoling Zhou, MD; Dingzhi Yue, MMed; Xu Liu, MMed; Nan Sun, MMed; Rong Hu, MMed; Xuesheng Li, MD; Fenglin Jia, MD; Gang Ning, MD; Haibo Qu[*], MD; Qiyuan Tian[*], PhD.

**Author Affiliations:** Department of Radiology, West China Second University Hospital, Sichuan University, Chengdu, China (Yijin Li, Yi Liao, Juncheng Zhu, Xuesheng Li, Fenglin Jia, Gang Ning, Haibo Qu); School of Biomedical Engineering, Tsinghua University, Beijing, China (Yijin Li, Mingxuan Liu, Le Chang, Hongjia Yang, Kasidit Anmahapong, Ziang Wang, Haoxiang Li, Xiaotian Hu, Yingqi Hao, Yifei Chen, Zihan Li, Qiyuan Tian); Oxford Centre for Integrative Neuroimaging, FMRIB, Nuffield Department of Clinical Neurosciences, University of Oxford, Oxford, UK (Ziyu Li); Department of Radiology, Sichuan Provincial Women's and Children's Hospital, The Affiliated Women's and Children's Hospital of Chengdu Medical College, Chengdu, China (Min Kang, Yan Song); Chengdu Women's and Children's Central Hospital, School of Medicine, University of Electronic Science and Technology of China, Chengdu, China (Hua Lai, Xiaoling Zhou); Yiliang County People's Hospital, Yiliang, China (Dingzhi Yue); The First People's Hospital of Yunnan Province, Kunming, China (Xu Liu); Department of Radiology, The Third Affiliated Hospital of Zhengzhou University, Zhengzhou, China (Nan Sun); Qujing Maternal and Child Health Hospital, Qujing, China (Rong Hu).

[#]These authors contributed equally.

[*]**Corresponding authors:** Qiyuan Tian, PhD, Center for Biomedical Imaging Research, Tsinghua University 30 Shuangqing Road, Haidian District, Beijing, China, PA100084 (qiyuantian@tsinghua.edu.cn). Haibo Qu, MD, No.20, Section 3, Renmin South Road, Chengdu, Sichuan Province, China, PA610041(windowsqhb@126.com).

**Summary**

**Background:** Reliable fetal brain biometry on magnetic resonance imaging (MRI) is essential for the diagnosis of congenital brain anomalies. Although artificial intelligence (AI) offers an automated approach, models trained on single-center data often degrade under cross-domain shifts, limiting their clinical translation. We aimed to develop a generalizable AI framework that delivers comprehensive, expert-level biometry across heterogeneous MRI sources.

**Methods:** In this retrospective, multicohort study, three-dimensional (3D) fetal brain volumes were reconstructed from thick-slice $T_2$-weighted MRI scans (gestational age 20–39 weeks) using slice-to-volume reconstruction, with 11 biometric measurements annotated by radiologists. We developed BioTTA, a backbone-agnostic, source-free unsupervised test-time adaptation (TTA) framework that adapts a pretrained model online to each individual unlabeled target volume under domain shifts. BioTTA was developed using turbo spin-echo (TSE) MRI data from West China Second University Hospital (WCSUH) and evaluated for biometry accuracy as the primary outcome (mean absolute error [MAE] and percentage of correct measurements within 4 mm [PCM@4 mm]) on three out-of-distribution (OoD) datasets representing cross-sequence (WCSUH-BTFE, i.e., balanced turbo-field-echo MRI data from WCSUH), cross-center (LFC from Shenzhen Bao'an Women's and Children's Hospital), and cross-nation (FeTA from University Children's Hospital Zurich) shifts. BioTTA was integrated into a web-based reporting platform that mapped measurements onto normative growth trajectories, with diagnostic performance compared with radiologists across four multi-center disease cohorts spanning three conditions: germinal matrix-intraventricular hemorrhage (WCSUH-TSE-GMH-IVH from WCSUH; OoD-GMH-IVH from Sichuan Provincial Women's and Children's Hospital and Chengdu Women's and Children's Central Hospital), ventriculomegaly (FeTA), and cerebellar abnormality (WCSUH-BTFE-CA from WCSUH).

**Results:** A total of 394 brain MRI volumes were analyzed. Across all three OoD datasets, BioTTA outperformed an existing pipeline (auto-SVRTK), the source-only model, and a domain-randomized model (FetalSynthGen), while matching the model finetuned with target labels. BioTTA achieved average MAEs of 1·460 ± 1·389 mm (FeTA), 1·152 ± 0·864 mm (WCSUH-BTFE), and 1·098 ± 1·200 mm (LFC) with the convolutional backbone (BioCNN), and 2·251 ± 1·844 mm (FeTA), 1·456 ± 1·140 mm (WCSUH-BTFE), and 1·866 ± 1·528 mm (LFC) with the transformer backbone (BioFormer).

Across the four disease cohorts, the platform achieved diagnostic accuracy comparable to that of a senior radiologist, ranging from 82·1% to 95·0%.

**Conclusion:** BioTTA delivers accurate, generalizable, expert-level biometry across heterogeneous MRI sources, enabling standardized prenatal brain growth assessment and scalable diagnosis of congenital brain anomalies across multicenter settings.

## 1. Introduction

Fetal brain anomalies severely compromise perinatal survival and lifelong neurodevelopment. Conditions such as ventriculomegaly (VM), germinal matrix-intraventricular hemorrhage (GMH-IVH), agenesis of the corpus callosum, and brain growth restriction affect 0·2–1·0% of pregnancies worldwide[1–4] and can lead to cerebral palsy, intellectual disability, epilepsy, or autism[3–5], placing substantial psychological and economic burdens on affected families. With declining fertility rates worldwide, ensuring optimal neurodevelopmental outcomes in every pregnancy is increasingly important, underscoring the need for systematic and standardized imaging assessment of the developing fetal brain[6,7].

Linear biometry is the most widely used and accessible quantitative approach for assessing fetal brain development. Recommended by the International Society of Ultrasound in Obstetrics and Gynecology (ISUOG)[8] as an integral component of prenatal screening, linear biometry is relevant to a global birth population of more than 100 million annually[6]. It quantifies the size of individual brain structures by measuring distances between predefined anatomical landmarks on standard imaging planes[9,10] and uses these measurements to assess abnormalities according to established diagnostic criteria. For example, a lateral ventricular atrial width of 10 mm or greater[11] defines VM, whereas a transverse cerebellar diameter (TCD) below the fifth percentile for gestational age (GA)[12] raises suspicion of cerebellar hypoplasia.

Current clinical imaging modalities often fail to provide the full set of predefined standard planes required for comprehensive linear biometry[13,14]. Specifically, two-dimensional (2D) ultrasound is constrained by skull ossification, fetal position, and maternal habitus, which can restrict acoustic access and limit visualization of deep midline and posterior fossa structures[15–17]. Conventional fetal magnetic resonance imaging (MRI) acquires the brain as separate 2D stacks of relatively thick slices and is susceptible to fetal motion. The resulting misoriented planes, incomplete plane coverage, and obscured anatomical landmarks due to partial-volume effects collectively compromise the accuracy and reproducibility of linear measurements[10,18]. These challenges can be addressed by using high-resolution isotropic brain volumes generated by fusing motion-corrected 2D MRI stacks through slice-to-volume reconstruction (SVR)[19–24].

However, manual biometry using 3D MRI volumes remains labor-intensive and expert-dependent[14]. It requires subjective slice selection and landmark placement, which not only are time-consuming, particularly when comprehensive measurements are required, but also contribute to substantial intra- and inter-observer variability, thereby compromising the efficiency, completeness, and consistency of assessment[25–27]. Recently, deep learning has enabled automated biometry, with segmentation-based[14,28] and heatmap-based landmark-detection[10,27,29–31] models approaching human-level accuracy in controlled settings.

Despite these advances, current models are not readily deployable in the real world due to limited generalizability. They are typically trained and evaluated on single-institution datasets and suffer from substantially degraded performance on out-of-distribution (OoD) data due to cross-domain discrepancies (Figure 1a, b), including differences in scanners, acquisition protocols, and SVR methods, as well as population heterogeneity in GAs and pathological conditions[23]. Existing strategies to improve robustness remain insufficient. Domain randomization fails to capture the complex and previously unseen shifts arising from imaging heterogeneity[32]. Supervised finetuning is constrained by the scarcity of expert annotations in target domains. Conventional unsupervised domain adaptation (UDA) requires access to source images during adaptation, which may be infeasible in multi-institutional fetal MRI because of privacy concerns[33].

To address the key barrier to the clinical translation of fetal brain biometry, we propose BioTTA, a backbone-agnostic, source-free unsupervised test-time adaptation (TTA) framework that enables expert-level and comprehensive measurements with cross-domain generalizability and scalable clinical applicability. Following supervised source-domain pretraining, BioTTA adapts online to each unlabeled target brain volume using length-consistency, entropy-minimization, and boundary-gradient losses together with atlas-guided anatomical priors, enabled by a differentiable sparse-weighted landmark detector that allows gradient propagation during adaptation (Figure 1c). We validated BioTTA for 11 guideline-recommended fetal brain biometrics (eFigure 1) across three OoD datasets representing cross-sequence, cross-center, and cross-nation shifts. BioTTA outperformed existing baselines and achieved performance comparable to supervised finetuning (Figure 1d, e). Moreover, BioTTA enabled expert-level diagnostic assessment across multiple disease cohorts by leveraging normative growth trajectories derived from radiologist-annotated data as a reference standard (Figure 1f). BioTTA is readily accessible via a web-based platform for clinicians and researchers (Figure 1g), paving the way for large-scale fetal neurodevelopmental studies and clinical applications that advance our understanding of human brain development in utero and improve prenatal care.

## 2. Methods

### 2.1 Study design and participants

This retrospective, multicohort, model development and validation study analyzed fetal brain $T_2$-weighted MRI data from five centers in China and Switzerland. In-house data were collected from West China Second University Hospital (WCSUH; September 2014–May 2024), Sichuan Provincial Women's and Children's Hospital (SPWCH; August 2025–September 2025), and Chengdu Women's and Children's Central Hospital (CWCCH; May 2024–September 2025). Two public datasets were also included: (i) the FeTA dataset from University Children's Hospital Zurich, and (ii) the LFC dataset from Shenzhen Bao'an Women's and Children's Hospital. Inclusion and exclusion criteria are provided in eAppendix 1. The study was approved by the institutional review boards of WCSUH (No. 2025-432), CWCCH (No. 2026-7-2), and SPWCH (No. 20250122-010), with informed consent waived owing to the retrospective use of de-identified data. The FeTA and LFC datasets were used under their respective data use agreements. This study is reported in accordance with the Standards for Reporting of Diagnostic Accuracy (STARD) 2015 guidelines[34] and adhered to the tenets of the Declaration of Helsinki.

The study followed a three-step design (eFigure 2), consisting of (i) supervised source-domain pretraining on the annotated turbo spin-echo data from WCSUH (WCSUH-TSE); (ii) biometric accuracy evaluation on three OoD datasets representing cross-sequence (WCSUH-BTFE, balanced turbo-field-echo data from WCSUH), cross-center (LFC), and cross-country (FeTA) shifts; and (iii) clinical diagnostic evaluation of the BioTTA-integrated reporting platform across four disease datasets covering germinal matrix-intraventricular hemorrhage (GMH-IVH; WCSUH-TSE-GMH-IVH and OoD-GMH-IVH from SPWCH and CWCCH), ventriculomegaly (VM; FeTA), and cerebellar abnormality (CA; WCSUH-BTFE-CA).

### 2.2 MRI acquisition, reconstruction, and preprocessing

In-house data were acquired as 2D thick-slice $T_2$-weighted image stacks in orthogonal orientations using TSE/FSE and BTFE sequences on 1·5T and 3T clinical scanners (GE, Siemens, and Philips) and reconstructed into 3D brain volumes at 0·8 mm isotropic resolution using NeSVoR[24] or NiftyMIC[35]. All volumes were processed using a uniform preprocessing pipeline comprising: (i) resampling to 0·8 mm isotropic resolution for anisotropic-resolution data (LFC and FeTA in this study), (ii) rigid registration to a standard spatiotemporal fetal brain atlas[36] space, (iii) cropping to 128 × 160 × 128 voxels, and (iv) voxel intensity normalization to [0, 1]. Detailed acquisition and reconstruction parameters are summarized in eTable 2.

### 2.3 Biometric annotation and reference standards

Following established fetal brain MRI biometric criteria[37], 11 linear biometric measurements were defined on standardized anatomical planes (eFigure 1): (i) right and left lateral ventricular widths in the axial (RLV-A, LLV-A) and coronal (RLV-C, LLV-C) planes; (ii) bone and cerebral biparietal diameters (BBD, CBD); (iii) transverse cerebellar diameter (TCD); (iv) fronto-occipital diameter (FOD); (v) anteroposterior vermis diameter (AVD); (vi) vermis height (VH); and (vii) corpus callosum length (CCL). Their definitions, clinical utility, and normative ranges are summarized in eTable 1. For model training and quantitative evaluation, ground-truth (GT) landmarks were annotated on the brain volumes with ITK-SNAP[38] (version 3.8). Specifically, initial landmark annotations across all datasets were performed by a junior radiologist (K.A.) and subsequently reviewed and corrected by a senior radiologist (F.J.; >10 years' experience in fetal neuroimaging) to establish the final reference standard. Reference diagnostic labels were assigned according to grading standards and expert review for the WCSUH-TSE-GMH-IVH, OoD-GMH-IVH, and FeTA datasets, and extracted from the original radiology reports for the WCSUH-BTFE-CA dataset (eAppendix 2).

### 2.4 BioTTA framework

BioTTA is a backbone-agnostic, source-free, unsupervised TTA framework that adapts a pretrained biometry model online to each unlabeled target volume during inference (eFigure 3). Specifically, during source-domain pretraining (Stage 1; eFigure 3a), BioTTA trains a biometry model comprising two branches: (i) a heatmap branch that predicts a spatial probability map for each landmark, and (ii) an auxiliary regression branch that estimates scalar biometric lengths via global pooling. BioTTA is compatible with different backbone architectures. In this study, the heatmap-localization backbone was instantiated using a convolutional network (BioCNN, based on 3D U-Net[39]) and a vision transformer (BioFormer, based on ViTPose[40]), both paired with the same CNN regression branch (eAppendix 3, eFigure 4). During testing (Stage 2; eFigure 3b), BioTTA updates only the affine parameters of the batch-normalization layers online following the TENT paradigm[41], while keeping all other model parameters frozen. BioTTA leverages three complementary self-supervised objectives to drive the adaptation, including: (i) a length-consistency loss that transfers domain-robust geometric information from the regression branch to the spatially sensitive heatmap branch, enabled by a differentiable sparse-weighted landmark detector that allows gradient propagation from measurement space to heatmaps; (ii) an entropy-minimization loss that sharpens localization confidence; and (iii) a boundary-gradient loss that anchors landmarks to high-contrast anatomical edges. Finally, BioTTA further leverages GA-matched spatiotemporal atlas priors to constrain landmark heatmaps and suppress anatomically implausible outliers. Detailed descriptions of the framework are provided in eAppendix 4.

**2.5 Comparison methods**

BioTTA was benchmarked against five representative methods spanning two categories (eAppendix 5). Target-label-free methods included (i) auto-SVRTK[10], a publicly available automated biometry and reporting pipeline representative of current open-source clinical tools; (ii) the source-only model, trained solely on the WCSUH-TSE source domain; and (iii) FetalSynthGen[32], a domain-randomization approach that augments source-domain training with synthetic intensity and contrast variations (eFigure 5). Target-supervised methods (finetuned with 10 expert-labeled target volumes) included (iv) DGMLD[29], a boundary-aware domain-generalization method tailored to fetal cerebellar biometry, and (v) supervised finetuning of the source-pretrained model, which represents the practical performance upper bound. Except for auto-SVRTK, all methods were evaluated using both BioCNN and BioFormer backbones. Additionally, component-wise ablation studies were performed on FeTA to investigate the contribution of each BioTTA component (eAppendix 6).

**2.6 Normative growth trajectories**

Normative growth trajectories for the Chinese population were constructed for all 11 biometrics over a GA range of 22–39 weeks from the manual annotations of normal fetuses (WCSUH-TSE dataset), following the parametric method of Royston and Wright[42]. Let $x$ denote GA and $y$ denote the biometric measurement. The expected mean trajectory, $\hat{\mu}(x)$, corresponding to the 50th centile, was modeled as a quadratic function of GA via least-squares regression ($\hat{\mu}(x) = c_0 + c_1 x + c_2 x^2$). The residuals were calculated as: $r_i = y_i - \hat{\mu}(x_i)$. Under the normal-residual assumption, scaled absolute residuals, $q_i = \sqrt{\pi/2}|r_i|$, were used to estimate the SD. To select the SD model, $q_i$ was linearly regressed against GA as $q_i = d_0 + d_1 x_i$, and the slope term $d_1$ was tested. A significant slope ($P < 0.05$) suggested GA-dependent variability, for which a linear GA-specific SD model, $\hat{\sigma}(x) = d_0 + d_1 x$, was used; otherwise, a constant SD model, $\hat{\sigma}(x) = \mathrm{SD}(r_i)$, was used. The 5th and 95th centiles were calculated as $\hat{\mu}(x) - 1.645\hat{\sigma}(x)$ and $\hat{\mu}(x) + 1.645\hat{\sigma}(x)$, respectively. In-sample Z-scores were calculated as $Z_i = [y_i - \hat{\mu}(x_i)]/\hat{\sigma}(x_i)$ and used to assess model adequacy in terms of approximate normality and reference-limit coverage. Approximate normality was evaluated using the Shapiro–Francia $W'$ test, whereas coverage was quantified as the proportion of Z-scores within $\pm 1.645$ and compared with the expected proportion of 90% using a two-sided exact binomial test.

**2.7 Clinical integration and reporting platform**

A web platform was developed to integrate BioTTA into an end-to-end automated reporting workflow (Figure 5). Given a reconstructed 3D brain volume in NIfTI format and the corresponding GA as inputs, the platform automatically performs preprocessing and BioTTA-based biometric prediction, maps each measurement onto the normative growth trajectories to derive GA-specific centiles, and outputs an interactive HTML dashboard with a WebGL-based 3D viewer, an exportable standardized PDF report, and a large language model (LLM)-generated diagnostic report. Implementation details are provided in eAppendix 7.

**2.8 Outcomes**

The primary outcome was biometric accuracy across the three OoD datasets. It was quantified using (i) the mean absolute error (MAE, mm) against expert reference measurements and (ii) the percentage of correct measurements (PCM), defined as the proportion of predicted measurements whose Euclidean deviation from the reference was below a prespecified threshold, for each of the 11 individual measurements and averaged across all measurements. A 4-mm deviation threshold (PCM@4 mm) was prespecified for clinical acceptance, consistent with previously reported manual annotation errors[27]. Secondary outcomes were (i) the diagnostic and grading performance of the platform (eAppendix 2) for three brain anomalies (i.e., GMH-IVH, VM, and CA), evaluated against assessments by seven radiologists (one senior radiologist, four attending radiologists, one junior radiologist, and one medical student) using accuracy, F1 score, and false-negative rate (FNR), and (ii) the agreement of biometric measurements from normal fetuses in OoD cohorts (FeTA, OoD-GMH-IVH, and WCSUH-BTFE-CA datasets; n = 121) with normative growth trajectories. Assessments for which radiologists were unable to determine a diagnosis or grade because of diagnostic uncertainty were excluded from the corresponding reader-specific denominators without imputation.

**2.9 Statistical analysis**

Measurement MAEs were reported as mean ± SD for the overall cohort and stratified by three GA intervals (< 29, 29–33, and > 33 weeks). Agreement between automated and reference measurements was assessed by Bland–Altman analysis with bias and 95% limits of agreement (LOA). Absolute errors were compared between BioTTA and each comparison method using a two-sided paired *t* test or a two-sided Wilcoxon signed-rank test when the paired differences were non-normal (Shapiro–Wilk $P < 0{\cdot}05$). To assess the biological plausibility of normative trajectory mapping, the one-sample Kolmogorov–Smirnov test was applied to determine whether GA-specific centiles derived from predictions in normal cases from OoD cohorts conformed to a standard uniform distribution. Because this test was repeated across the 11 biometrics, raw *P* values ($P_{raw}$) were adjusted using the Benjamini–Hochberg procedure to control the false discovery rate, and both $P_{raw}$ and adjusted values ($P_{BH}$) were reported. Statistical significance was defined as a two-sided $P < 0{\cdot}05$. Analyses were performed using Python (SciPy 1.15.1 and scikit-learn 1.3.2).

## 3. Results

The final study cohort consisted of 394 fetal brain MRI volumes from five centers in China and Switzerland (GA 20–39 weeks) after exclusions (eFigure 2, eTable 2). The source-domain WCSUH-TSE cohort included 66 normal brain volumes. For OoD biometry evaluation, three cohorts (WCSUH-BTFE, LFC, and FeTA) each provided 40 test volumes and 10 finetuning volumes. For diagnostic performance evaluation, four cohorts (WCSUH-TSE-GMH-IVH, OoD-GMH-IVH, FeTA, and WCSUH-BTFE-CA) included 39, 77, 40, and 62 cases, respectively, comprising 39 GMH-IVH cases, 65 normal and 12 GMH-IVH cases, 19 normal and 21 VM cases, and 37 normal and 25 CA cases.

Across both backbones and all three OoD datasets, BioTTA consistently matched supervised finetuning in terms of the average MAE across 11 biometrics ($P > 0{\cdot}05$) while outperforming all other comparison methods ($P < 0{\cdot}05$). Using the BioCNN backbone (Figure 3a, eTable 3), BioTTA significantly reduced the average MAE on FeTA by 47·1% relative to the source model (from 2·761 ± 5·029 mm to 1·460 ± 1·389 mm; $P < 0{\cdot}05$), matching the supervised finetuned model (1·495 ± 1·788 mm; $P > 0{\cdot}05$). Its average MAE on WCSUH-BTFE (1·152 ± 0·864 mm) and LFC (1·098 ± 1·200 mm) were likewise comparable to those of the supervised finetuned model (1·199 ± 0·922 mm and 1·097 ± 1·101 mm, respectively; all $P > 0{\cdot}05$). Consistent with these improvements, BioTTA's PCM@4 mm reached 94·8% on FeTA, 98·8% on WCSUH-BTFE, and 96·7% on LFC (vs 87·7%, 96·3%, and 88·3% for the source model), indicating that most automated measurements fell within the clinically acceptable error margin. By contrast, competing methods showed limited robustness under domain shift: (i) auto-SVRTK achieved performance comparable to the source model, suggesting limited generalizability; (ii) FetalSynthGen substantially improved performance only on WCSUH-BTFE; and (iii) DGMLD showed degraded generalization on LFC, resulting in a higher average MAE than the source model (7.261 ± 3.968 mm vs 2.046 ± 2.916 mm).

Using the BioFormer backbone (Figure 3b, eTable 4), BioTTA achieved even greater reductions in average MAE relative to the source model, with decreases of 55·7% on FeTA (from 5·080 ± 6·986 mm to 2·251 ± 1·844 mm), 51·8% on WCSUH-BTFE (from 3·020 ± 4·975 mm to 1·456 ± 1·140 mm), and 82·0% on LFC (from 10·344 ± 11·103 mm to 1·866 ± 1·528 mm). BioTTA did not differ significantly from the supervised finetuned model on FeTA, WCSUH-BTFE, or LFC (2·948 ± 4·508 mm, 1·739 ± 2·081 mm, and 2·507 ± 3·036 mm, respectively; all $P > 0·05$), despite achieving numerically lower errors. Consistent with these improvements, BioTTA's PCM@4 mm reached 84·1% on FeTA, 96·1% on WCSUH-BTFE, and 91·7% on LFC (vs 67·1%, 87·0%, and 49·2% for the source model), exceeding the corresponding values of the finetuned model (82·1%, 93·9%, and 85·0%). Consistent with the BioCNN results, FetalSynthGen substantially improved performance only on LFC, whereas DGMLD performed substantially worse than BioTTA on LFC, with a higher average MAE (6·589 ± 4·333 mm).

Moreover, BioTTA showed minimal measurement bias (Figure 3c, d). With the BioCNN backbone, BioTTA achieved an overall bias of −0·043 mm with narrow LOA ([−3·539, 3·452] mm), and maintained individual biometric biases close to the zero-difference line (−0·912 to 0·809 mm). With the BioFormer backbone, BioTTA showed similar agreement (overall bias, −0·111 mm; LOA, [−5·001, 4·780] mm), confirming consistent performance across backbone architectures. Together, these findings indicate that BioTTA provides reliable and unbiased measurements for mapping onto normative growth trajectories and estimating GA-specific centiles.

Qualitative assessment of representative cases was consistent with the quantitative findings (Figure 2). Across FeTA, WCSUH-BTFE, and LFC, the source-only model exhibited substantial domain shift, with landmarks displaced from anatomical boundaries and predicted lengths deviating markedly from the GT for several measurements (e.g., TCD and AVD in FeTA case 1; LLV-A, LLV-C, FOD, AVD, VH, and CCL in FeTA case 2). FetalSynthGen recovered some measurements (e.g., TCD in FeTA case 1), but failed on others (e.g., FOD in WCSUH-BTFE case 1) and occasionally worsened measurements relative to the source model (e.g., AVD and VH in LFC case 1). In contrast, BioTTA consistently localized landmarks close to the GT and produced near-reference measurements across anatomical planes and GAs, visually approaching the supervised finetuned model.

BioTTA provided consistent improvements across fetal development (eFigure 8). Specifically, it reduced errors across all GA intervals and both backbones, with the greatest improvements observed at earlier GAs (< 29 weeks). Notably, with the BioFormer backbone, BioTTA achieved significant improvements across nearly all GA strata and in the pooled cohort ($P < 0·001$), confirming robust generalization across the clinically relevant gestational window.

BioTTA achieved improvements through both the self-supervised adaptation losses and the atlas prior, as shown by component-wise ablation on FeTA (eTable 5, eFigure 6). Relative to the source model, applying the length-consistency loss alone lowered the average MAE from 2·761 ± 5·029 mm to 1·896 ± 2·941 mm (31·3% reduction), whereas applying the entropy-minimization or boundary-gradient loss alone increased the average MAE to 5·948 ± 10·838 mm (115·4% increase) and 3·016 ± 5·795 mm (9·2% increase), respectively. Combining all three self-supervised losses lowered the average MAE to 1·635 ± 1·709 mm (40·8% reduction), while applying atlas guidance alone reduced it to 1·995 ± 2·483 mm (27·7% reduction). Incorporating all components, the full BioTTA achieved the lowest average MAE of 1·460 ± 1·389 mm (47·1% reduction), significantly outperforming all ablation configurations (all P < 0·05). Mechanistically, BioTTA realigned target-domain features toward the source domain during adaptation, as shown by *t*-SNE visualization (eFigure 7).

Normative growth trajectories for 11 fetal brain biometrics were established across 22–39 weeks of GA using the Chinese WCSUH-TSE cohort (Figure 4, eTable 6, eTable 7, eTable 8). First, all fitted trajectories were statistically adequate across all 11 biometrics (eTable 8), with Z-score distributions approximating Gaussian distributions for 10 of the 11 biometrics (all except TCD; Shapiro–Francia $W'$ 0·971–0·990, all $P > 0·05$) and reference-limit coverage matching the expected 90% (86·36–93·94% within $\pm 1·645$ SD; all $P \geq 0·304$). Second, the derived trajectories differed from established Western MRI-based reference standards (Tilea and colleagues[43] for CCL; Kyriakopoulou and colleagues[44] for the remaining 10 biometrics; Figure 4a), with ventricular measurements (RLV-A, LLV-A, RLV-C, and LLV-C) consistently above the corresponding Western references; BBD, FOD, vermis AVD and VH, and CCL slightly below; and TCD and CBD closely aligned. Third, the trajectories generalized beyond the source domain (Figure 4b). Most BioTTA predictions for the 121 OoD normal fetuses fell within the 5th–95th centile range, and GA-conditioned centile distributions were consistent with uniformity for 10 of the 11 biometrics after false-discovery-rate correction (Kolmogorov–Smirnov $P_{BH} \geq 0·102$), with only LLV-C showing significant deviation ($P_{BH} < 0·001$). Finally, the trajectories captured abnormal growth patterns across all four disease cohorts (Figure 4c–f). RLV-C and LLV-C widths exceeded the upper centiles in fetuses with Grade II/III GMH-IVH (WCSUH-TSE-GMH-IVH and OoD-GMH-IVH) or VM (FeTA), whereas TCD, AVD, and VH values fell below the lower centiles in fetuses with CA (WCSUH-BTFE-CA). Together, these findings demonstrate that the normative trajectories provide a population-specific reference framework for fetal neurodevelopmental assessment and can support integration into clinical reporting workflows.

Based on the normative trajectories, the web-based reporting platform converted each BioTTA-derived measurement into a GA-specific centile and generated a structured diagnostic report (Figure 5a-b, eAppendix 7). For example, the platform localized all 11 landmarks on the reconstructed volume of a representative normal fetus (GA 31.0 weeks, FeTA), plotted each measurement against the normative curves (Figure 5c), and generated a structured LLM-based report indicating no abnormal biometric findings (eFigure 12). In three representative disease cases, the platform identified biometrics that deviated from the reference range: (i) LLV-A and LLV-C widths above the 95th centile in a fetus with GMH-IVH (GA 30.1 weeks, OoD-GMH-IVH; eFigure 9); (ii) RLV-A and RLV-C widths above the 95th centile in a fetus with VM (GA 23.4 weeks, FeTA; eFigure 10); and (iii) TCD and VH below the 5th centile in a fetus with CA (GA 25.7 weeks, WCSUH-BTFE-CA; eFigure 11). These abnormalities were summarized in the LLM-generated reports, which provided case-level risk stratification with corresponding follow-up suggestions (eFigures 13–15).

The BioTTA-integrated platform demonstrated expert-level performance across all four disease cohorts (Figure 5d, eTable 9). For binary anomaly detection, the platform achieved accuracy comparable to that of a senior radiologist for VM detection in the two GMH-IVH cohorts (82·1% vs 82·1% in WCSUH-TSE-GMH-IVH; 93·5% vs 94·8% in OoD-GMH-IVH), and higher accuracy for VM detection in FeTA (95·0% vs 92·5%) and CA detection in WCSUH-BTFE-CA (90·3% vs 82·3%). The platform also achieved lower FNRs than the senior radiologist for VM detection in OoD-GMH-IVH (0·200 vs 0·300) and CA detection in WCSUH-BTFE-CA (0·240 vs 0·320). Its largest advantage was observed in CA subtyping, where it outperformed the senior radiologist in detecting vermian agenesis (accuracy: 95·2% vs 75·8%; F1 score: 0·903 vs 0·118; FNR: 0·067 vs 0·933) and cerebellar hypoplasia (accuracy: 90·3% vs 74·2%; F1 score: 0·750 vs 0·529). By contrast, the platform did not surpass the senior radiologist in finer grading tasks, including GMH-IVH grading (66·7% vs 71·8% in WCSUH-TSE-GMH-IVH; 90·9% vs 94·8% in OoD-GMH-IVH) and grading of right lateral VM severity in FeTA (62·5% vs 75·0%). Nevertheless, its performance remained within the accuracy range of the four attending radiologists for these grading tasks (53·8–71·8% in WCSUH-TSE-GMH-IVH; 87·0–93·5% in OoD-GMH-IVH; 50·0–75·0% in FeTA).

## 4. Discussion

To our knowledge, this multicohort study is the first to develop and validate a backbone-agnostic, source-free unsupervised TTA framework, BioTTA, that enables robust fetal brain biometry across previously unseen MRI sequences, centers, and countries without site-specific retraining. Across three OoD datasets and two network backbones, BioTTA reduced average measurement errors by 46–82% relative to the source model and achieved comparable performance to supervised target-domain finetuning. BioTTA also produced measurements with minimal systematic bias, provided consistent improvements across GAs, and reliably mapped OoD cases onto normative growth trajectories for 11 guideline-recommended fetal brain biometrics. When integrated into a web-based platform, BioTTA facilitated the detection of multiple fetal brain abnormalities across four disease cohorts, with diagnostic performance comparable to that of a senior radiologist. Together, these findings establish BioTTA as a practical framework for standardized and scalable fetal brain biometry, facilitating the clinical translation of automated prenatal neurodevelopmental assessment across heterogeneous imaging environments.

BioTTA achieves robust cross-domain generalizability by regularizing image appearance adaptation at test time with anatomy-guided geometric constraints. Image appearance shifts between the source and target domains (e.g., tissue-intensity distributions, tissue contrast, and boundary sharpness) can bias landmark localization and consequently degrade biometry accuracy[45]. BioTTA addresses these shifts by combining the entropy-minimization loss, a widely adopted objective in conventional TTA for image segmentation, with the boundary-gradient loss specially designed in this study for landmark localization. Specifically, the entropy-minimization loss sharpens landmark heatmaps to increase prediction confidence but does not guarantee that the resulting landmark locations are anatomically plausible. The boundary-gradient loss encourages landmarks to align with high-contrast image edges but does not guarantee that these edges correspond to the intended anatomical structures. Therefore, using either loss alone did not reduce the average MAE of biometric measurements relative to the source model (5·948 mm and 3·016 mm vs 2·761 mm, respectively). BioTTA incorporates anatomy-guided geometric constraints to address this issue. Specifically, it uses the GA-matched atlas prior to anchor absolute landmark positions and the inter-landmark length-consistency loss to constrain relative landmark relationships. When applied individually, these two geometric constraints reduced the average MAE from 2·761 mm to 1·995 mm (27·7% reduction) and 1·896 mm (31·3% reduction), respectively. When applied jointly, these two constraints stabilized the image appearance adaptation process, enabling BioTTA to achieve the lowest average MAE of 1·460 mm (47·1% reduction) and significantly outperform all ablation configurations (all $P < 0{\cdot}05$).

BioTTA's strong generalizability offers substantial benefits for clinical deployment and broader clinical translation. By performing source-free, sample-wise adaptation to each unlabeled OoD scan without prior knowledge of the target-domain distribution, BioTTA can accommodate diverse and previously unseen shifts across sequence types (e.g., TSE/FSE and BTFE), field strengths (e.g., emerging 0·55T[10], 1·5T, and 3T), scanner vendors (e.g., United Imaging, GE, Siemens, and Philips), and SVR methods (e.g., NeSVoR, NiftyMIC, and SVRTK), all of which may introduce small but systematic differences in brain biometry[23]. This generalizability could support more consistent interpretation of fetal brain abnormalities across institutions and broaden access to expert-level prenatal neurodevelopmental assessment, particularly in regions where specialized fetal neuroradiology expertise remains limited[46]. Crucially, because BioTTA requires neither additional target-domain retraining nor target-domain data collection and annotation, it can be readily integrated into a web-based platform for rapid and cost-effective deployment with reduced computational and infrastructure requirements, further enhancing its clinical usability and accessibility.

Another key advantage of BioTTA is its capacity for comprehensive, multi-structure fetal brain phenotyping. BioTTA simultaneously quantifies 11 guideline-recommended biometrics within a unified framework, enabling coordinated assessment of abnormalities across multiple brain structures. For example, by integrating TCD, AVD, and VH relative to GA-specific centiles, BioTTA achieved an accuracy of 90·3% for CA detection, exceeding that of the senior radiologist (82·3%). By incorporating these cerebellar and vermian measurements, BioTTA may also provide quantitative information relevant to the clinical differentiation of posterior-fossa abnormalities, including Blake pouch cyst, vermian hypoplasia, and Dandy–Walker malformation[47,48]. Additionally, by jointly assessing cranial (e.g., BBD and CBD), ventricular (e.g., RLV-A and LLV-A), and posterior-fossa (e.g., TCD and VH) measurements, BioTTA may facilitate recognition of intracranial patterns associated with open spina bifida, including reduced biparietal and cerebellar dimensions and VM[49].

BioTTA's combined generalizability and comprehensiveness make the construction of normative fetal brain growth charts feasible at a scale unattainable by manual annotation. Specifically, BioTTA's negligible systematic bias in OoD cohorts (−0·043 mm for BioCNN and −0·111 mm for BioFormer) supports the reliable aggregation of measurements from numerous heterogeneous sources into large population-level cohorts for deriving population-specific growth trajectories. These trajectories themselves constitute quantitative descriptions of fetal brain development, enabling investigation of population-specific developmental patterns across geographic and ancestral populations. For example, the Chinese trajectories derived in this study exceeded Western references for ventricular widths but fell below them for BBD, FOD, AVD, VH, and CCL[43,44]. These population-specific reference trajectories also provide a normative framework for quantifying deviations from expected brain development, which could inform clinical decision-making for abnormalities defined or characterized by biometric thresholds and enable quantification of structure-specific and GA-dependent brain-growth alterations associated with conditions such as congenital heart disease[50], fetal growth restriction[51], and other understudied disorders.

Several limitations of the current work warrant consideration. First, although this study included multi-center data for model development and external evaluation, the validation was retrospective. Evaluating BioTTA prospectively within routine clinical workflows will further validate its performance under real-world clinical conditions and assess its readiness for clinical implementation. Second, this study used data collected from five centers in China and Switzerland. Expanding validation to broader geographic and demographic settings will further establish BioTTA's generalizability across diverse populations. Third, this study included relatively small disease cohorts. Expanding evaluations to larger, prospectively assembled cohorts will further validate BioTTA's diagnostic performance. Fourth, this study did not evaluate LLM-generated reports through reader studies. Comprehensively evaluating their factual accuracy, readability, and clinical utility across readers with different levels of expertise will help optimize the reporting module for clinical use.

In summary, BioTTA marks a shift in fetal brain MRI analysis, transforming domain-specific automated biometry into accurate, anatomically plausible, and clinically applicable assessment across heterogeneous imaging sources. By overcoming the key translational barrier of cross-domain performance degradation without requiring target annotations or access to source data, it supports reliable normative growth evaluation and expert-level diagnosis of congenital brain abnormalities across multicenter settings. This capability lays the groundwork for prospective, broader geographic studies aimed at standardizing fetal neurodevelopmental assessment and integrating adaptive artificial intelligence into routine prenatal imaging practice.

**Data sharing statement**

The raw fetal brain MRI data cannot be shared publicly because of patient privacy and ethical considerations. Public fetal brain MRI datasets are available from the FeTA dataset (https://fetachallenge.github.io/pages/Data_download) and the LFC dataset (https://drive.google.com/drive/folders/1PVOi5dX4SSjbZ4TqjRPZhfewj7fgFY9p; https://github.com/lhaof/LFC). Source code is publicly available for BioTTA (https://github.com/birthlab/BioTTA), auto-SVRTK (https://github.com/SVRTK/auto-proc-svrtk), DGMLD (https://github.com/lhaof/DGMLD), and ViTPose (https://github.com/vitae-transformer/vitpose).

**Declaration of interests**

All authors declare no competing interests.

**Contributors**

Y.Li: Conceptualization, Methodology, Software, Investigation, Formal analysis, Writing – original draft, Visualization. M.L.: Conceptualization, Methodology, Software, Investigation, Formal analysis, Writing – original draft, Visualization. Y.Liao: Data curation, Validation, Investigation, Writing – original draft. L.C.: Methodology, Software, Validation, Investigation, Writing – original draft. H.Y.: Investigation, Formal analysis, Software, Writing – review & editing. K.A.: Investigation, Software. X.H.: Software, Investigation, Visualization. Z.W.: Software, Investigation. Y.H.: Methodology, Software, Investigation, Formal analysis. H.Li: Data curation, Investigation. Y.C.: Investigation, Software. ZH.L.: Writing – review & editing. ZY.L.: Writing – review & editing. J.Zhu: Data curation, Methodology, Validation, Software. M.K.: Resources, Data curation. Y.S.: Resources, Data curation. H.Lai: Resources, Data curation. X.Z.: Resources, Data curation. D.Y.: Investigation, Validation. X.Liu: Investigation, Validation. N.S.: Investigation, Validation. R.H.: Investigation, Validation. X.Li: Investigation, Validation. F.J.: Investigation, Validation. G.N.: Supervision, Project administration, Resources. H.Q.: Conceptualization, Supervision, Project administration, Funding acquisition, Resources, Writing – review & editing. Q.T.: Conceptualization, Methodology, Supervision, Project administration, Funding acquisition, Writing – review & editing.

**Acknowledgements**

This work was supported by the Natural Science Foundation of Sichuan Province (grant number 2025ZNSFSC1768), the Scientific Research Project of Sichuan Medical Association (grant number 2024HR130), the Science and Technology Department of Sichuan Province, China (grant number 25SYSX0255), the Tsinghua University Startup Fund, and the Tsinghua University Dushi Program (grant numbers 20241080026, 20251080056, 20261080018).

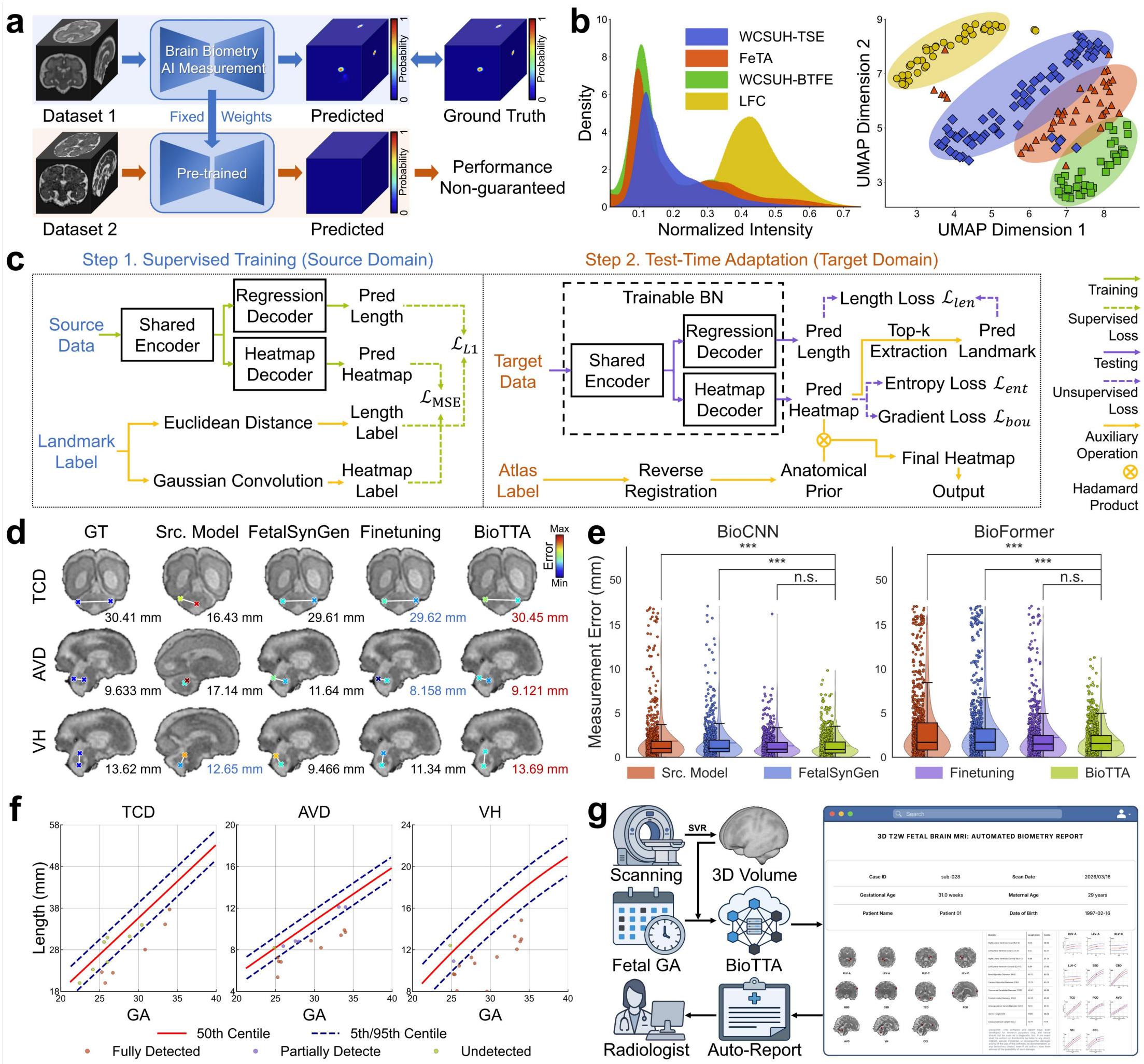

**Figure 1. Study Overview.** BioTTA is a source-free, unsupervised test-time adaptation (TTA) framework that enables generalizable, expert-level fetal brain biometry across heterogeneous MRI data. **a-b**, The challenge of domain shift in automated fetal brain biometry. Applying source-trained models to data from different domains leads to substantial performance degradation. Feature distribution analysis via uniform manifold approximation and projection (UMAP) reveals a pronounced domain gap between datasets. **c,** Overview of the BioTTA framework, which consists of two steps. In Step 1, a backbone-agnostic network (BioCNN or BioFormer) comprising a shared encoder, a regression decoder, and a heatmap decoder is trained on manually annotated source-domain data to jointly predict landmark heatmaps and biometric measurements. In Step 2, the pretrained network is adapted at test time to each unlabeled out-of-distribution (OoD) target sample through self-supervised optimization, leveraging length-consistency, entropy-minimization, and boundary-gradient losses. Atlas-guided anatomical priors are further incorporated to enforce geometric plausibility. **d,** Visualization of cerebellar biometric measurements (TCD, AVD, and VH) in a representative target-domain case from LFC. **e,** Distribution of measurement errors across methods, showing no significant differences between BioTTA and supervised target-domain finetuning for either backbone ($P > 0.05$). **f,** Selected normative growth trajectories for the Chinese population established in this study, overlaid with BioTTA-derived measurements from OoD cases with cerebellar abnormalities (red, fully detected; purple, partially detected; green, undetected). **g,** Clinical integration. A web-based system powered by BioTTA automatically generates fetal brain biometry reports to assist radiologists with interpretation.

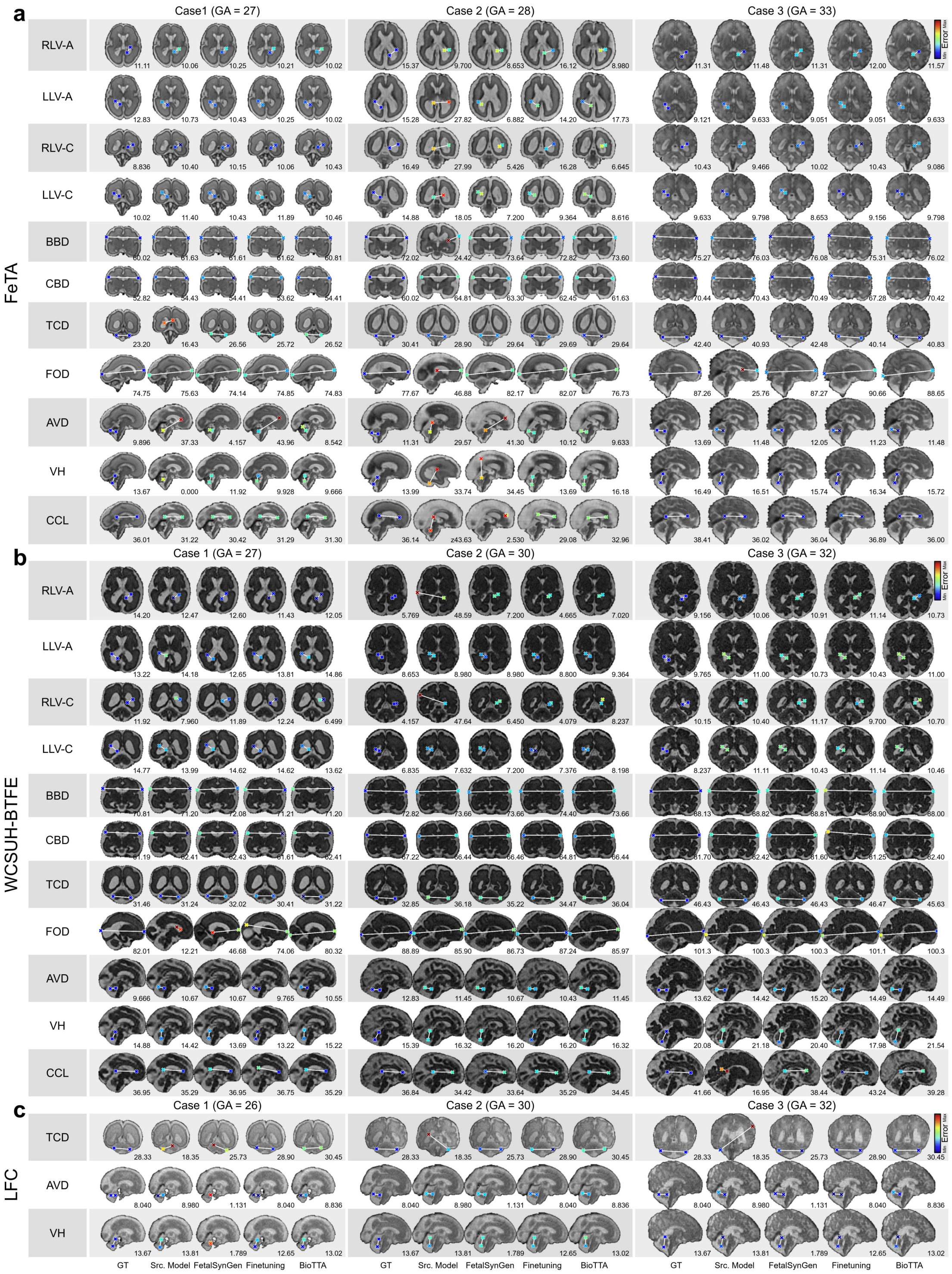

**Figure 2. Visualization of Biometric Measurement Results across Different Methods.** Three representative cases from the FeTA (**a**), WCSUH-BTFE (**b**), and LFC (**c**) datasets are visualized across gestational ages (GAs) from 26 to 33 weeks. The proposed BioTTA is compared with the source model, FetalSynthGen, and a finetuning method with a BioCNN backbone. Eleven biometric measurements are evaluated using ground truth (GT) as the reference, though only three are displayed for the LFC dataset (**c**) due to limited public GT annotations. The landmark localization error is indicated by the color of the two landmarks defining each measurement, with the color representing the Euclidean distance from the corresponding GT coordinates. The predicted measurement lengths (in mm) are annotated for all cases.

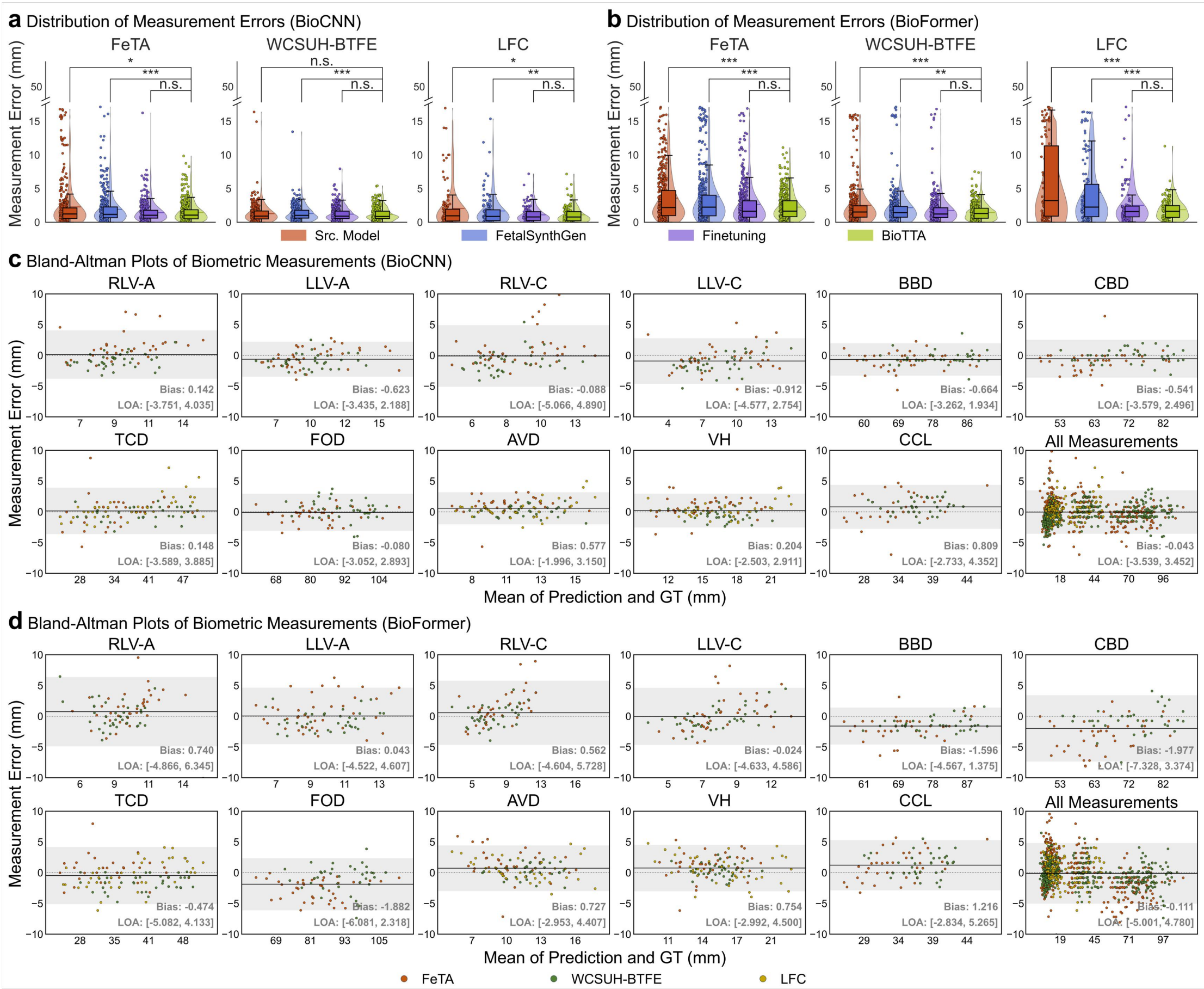

**Figure 3. Quantitative Evaluation of Biometric Predictions on the Target Domains. a, b,** Raincloud plots visualize the statistical distributions of length measurement errors across the BioCNN (**a**) and BioFormer (**b**) backbones on three target datasets (FeTA, WCSUH-BTFE, and LFC; $n = 40$ per dataset). The proposed BioTTA is compared with the source model, FetalSynthGen, and a finetuning method. Statistical significance of the differences is annotated for each comparison (n.s., $P > 0.05$; *, $P \leq 0.05$; **, $P \leq 0.01$; ***, $P \leq 0.001$). **c, d,** Bland-Altman plots of 11 predicted biometric measurements quantify the systematic deviation of BioCNN (**c**) and BioFormer (**d**) predictions from the ground truth across all three datasets (total $n$ = 120 cases). Annotated values indicate the measurement bias (mean difference, solid line) and the limits of agreement (LOA), highlighted by the shaded region (bias ± 1.96 standard deviations). The dashed line represents the ideal zero-difference baseline.

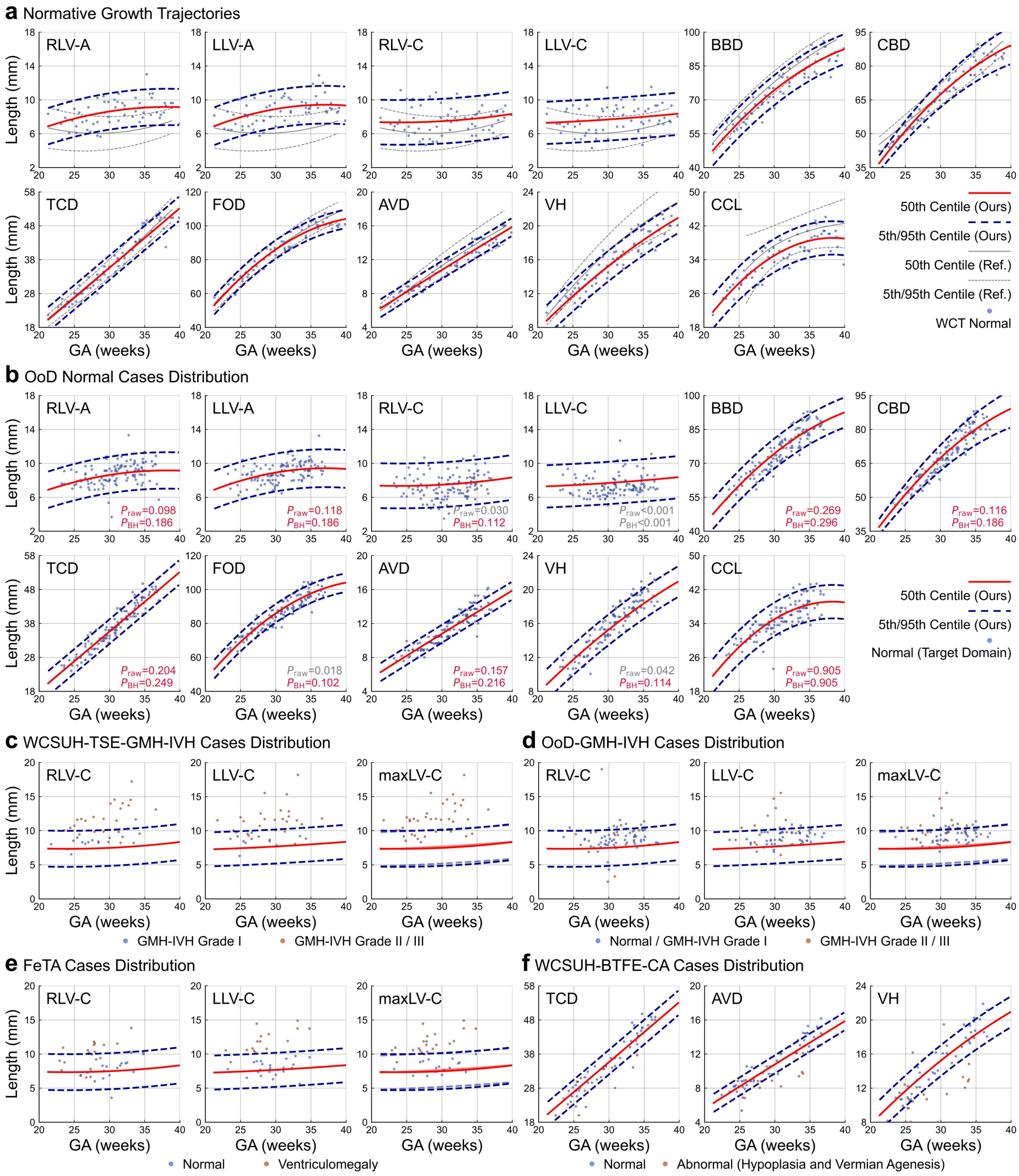

**Figure 4. Derivation and Clinical Evaluation of Normative Fetal Growth Trajectories. a,** Normative growth curves for fetal biometry across gestational age (GA) are derived from the source training dataset ($n$ = 66). Scatter points represent the expert annotations used for model training. To validate the derived trajectories, our estimated 50th percentile (median, red solid line) and reference intervals (5th and 95th percentiles, dark-blue dashed lines) are compared against an established external reference standard (corresponding black solid and dashed lines). **b,** Validation of normative growth trajectories on healthy target datasets. Biometric measurements estimated by BioTTA from normal cases ($n$ = 121) across three target datasets (FeTA, OoD-GMH-IVH, and WCSUH-BTFE-CA) are plotted against the normative growth trajectories derived from the source (WCSUH-TSE). Gestational age-conditioned centiles were calculated from the source-derived reference distribution and assessed for uniformity using a one-sample Kolmogorov–Smirnov test. Raw $P$ values ($P_{\text{raw}}$) and Benjamini–Hochberg-adjusted $P$ values ($P_{\text{BH}}$) are calculated and annotated for each biometry, with $P_{\text{BH}} < 0.05$ indicating significant departure from the normative reference after controlling for the false discovery rate. **c–f,** Clinical evaluation. BioTTA-predicted measurements are plotted against the established normative baseline to assess growth deviations across four target datasets: the WCSUH-TSE-GMH-IVH (**c**) and OoD-GMH-IVH (**d**) datasets for GMH-IVH diagnosis, the FeTA dataset (**e**) for ventriculomegaly (VM) diagnosis, and the WCSUH-BTFE-CA dataset (**f**) for cerebellar abnormality (hypoplasia and vermian agenesis) diagnosis. Scatter point colors encode the reference diagnostic labels established via expert measurement annotations.

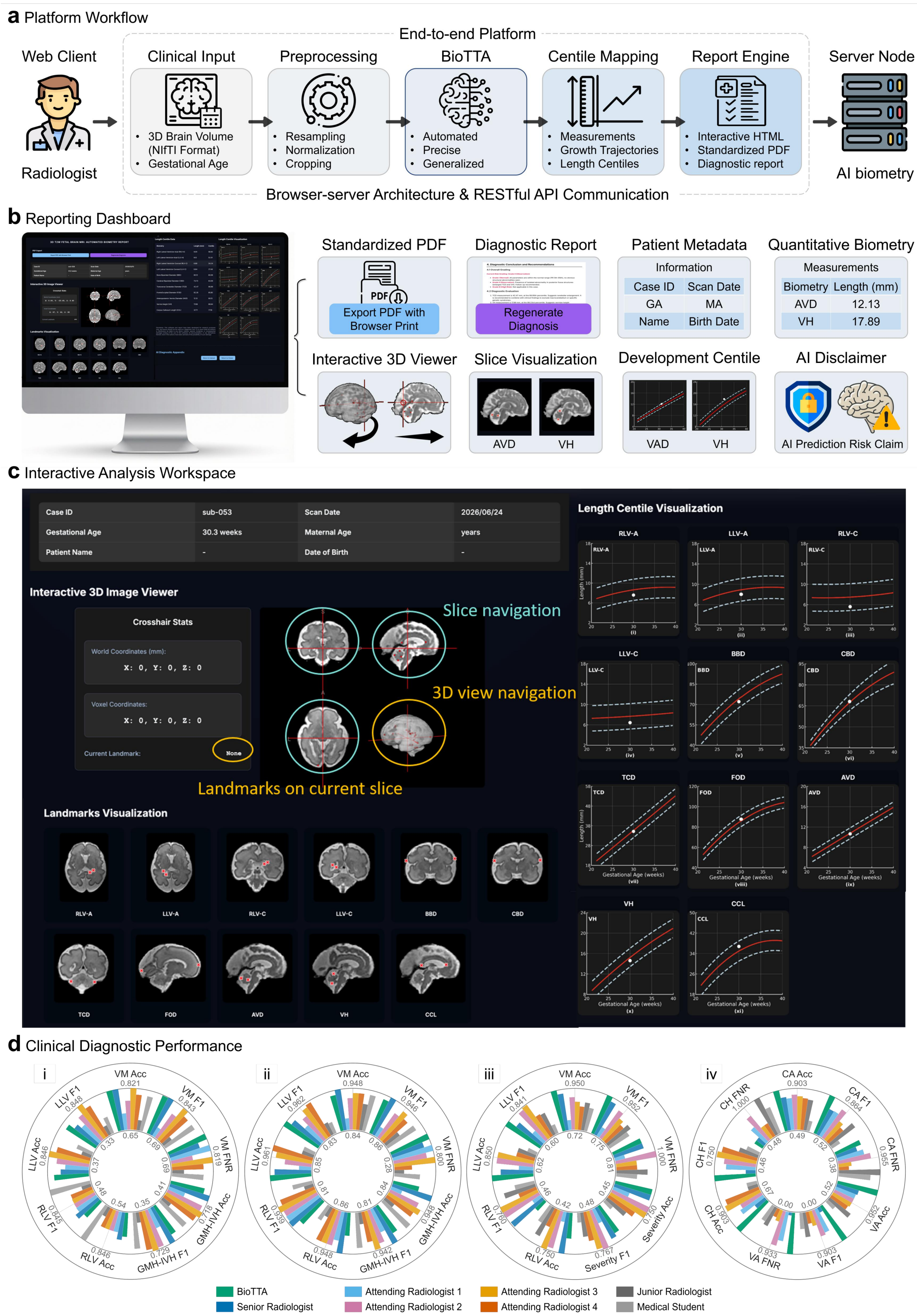

**Figure 5. Workflow, Interactive Dashboard and Diagnostic Performance of the Reporting Platform. a**, End-to-end platform workflow. Radiologists upload 3D brain volumes in NIfTI format reconstructed using NiftyMIC or NeSVoR and corresponding gestational age (GA) via a web client. The input data are automatically preprocessed and analyzed by the BioTTA framework, which ensures robust cross-domain biometry via test-time adaptation. Predicted measurements are then mapped to normative growth trajectories to calculate length centiles. Operating on a browser-server architecture with RESTful APIs, the report engine compiles these outputs into an interactive HTML dashboard, an exportable PDF, and an LLM-generated diagnostic report. **b**, Interactive reporting dashboard. The browser-based interface

displays essential patient metadata alongside comprehensive analytical outputs. Visual and quantitative modules include an interactive 3D viewer for volumetric inspection, multi-planar reconstructions of standard anatomical views, a detailed table of biometric measurements (in mm), and patient-specific growth curves. An LLM-generated diagnostic report and a standardized PDF report are provided with one-click regeneration options. To ensure safe and transparent clinical translation, all AI-generated diagnostic outputs are accompanied by explicit diagnostic disclaimers. **c**, Web-based reporting platform output for a representative normal fetus (GA 31.0 weeks, FeTA). All biometric measurements align with normative trajectories (50th percentile, red solid line; 5th and 95th percentiles, blue dashed lines), falling within expected reference ranges and supporting normal development. The corresponding diagnostic report is shown in eFigure 12. **d**, Clinical diagnostic performance of the reporting platform. Radar plots compare the platform with seven readers of descending seniority (senior radiologist, attending radiologists 1-4, junior radiologist, and medical student) across four OoD cohorts, including WCSUH-TSE-GMH-IVH (**i**) and OoD-GMH-IVH (**ii**) for germinal matrix hemorrhage-intraventricular hemorrhage (GMH-IVH), FeTA (**iii**) for ventriculomegaly (VM), and WCSUH-BTFE-CA (**iv**) for cerebellar abnormality (CA). Diagnoses were derived from BioTTA-predicted measurements mapped to gestational-age-specific normative thresholds following established clinical criteria, with full numerical results in eTable 9.